\documentclass[10pt,letterpaper]{article}
\usepackage[top=0.85in,left=2.75in,footskip=0.75in]{geometry}
\usepackage{amsmath,amssymb}
\usepackage{epstopdf}
\usepackage{graphicx}
\usepackage{changepage}

\usepackage{textcomp,marvosym}

\usepackage{cite}

\usepackage{nameref,hyperref}

\usepackage[right]{lineno}

\usepackage[nopatch=eqnum]{microtype}
\DisableLigatures[f]{encoding = *, family = * }

\usepackage[table]{xcolor}

\usepackage{array}

\usepackage{multirow}

\usepackage{adjustbox}
\newcolumntype{+}{!{\vrule width 2pt}}

\newlength\savedwidth

\raggedright
\usepackage[aboveskip=1pt,labelfont=bf,labelsep=period,justification=raggedright,singlelinecheck=off]{caption}

\makeatletter
\renewcommand{\@biblabel}[1]{\quad#1.}
\makeatother

\usepackage{lastpage,fancyhdr,graphicx}
\usepackage{epstopdf}

\fancyheadoffset[L]{2.25in}
\fancyfootoffset[L]{2.25in}
\begin{document}
\vspace*{0.2in}

\begin{flushleft}
{\Large
\textbf{Assisting Patients with Reliable Information Gathering: A Guide to Building Efficient QA Systems for Medical Use} 
}
\newline
\\
Zhipeng Yin\textsuperscript{1},
Zichong Wang\textsuperscript{1},
Min Chen\textsuperscript{1},
Ian Stockwell \textsuperscript{2},
Xin Ning\textsuperscript{1},
Jun Liu\textsuperscript{3*},
Wenbin Zhang\textsuperscript{1*}
\\
\bigskip
\textbf{1} Florida International University, Miami, Florida, United States of America\\
\textbf{2} University of Maryland Baltimore County, Baltimore, Maryland, United States of America\\
\textbf{3} Carnegie Mellon University, Pittsburgh, Pennsylvania, United States of America


\bigskip

%
%





* Corresponding author: liujun@cmu.edu, wenbin.zhang@fiu.edu

\end{flushleft}


%
\section*{Abstract}
The application of question-answering (QA) systems in the medical domain has rapidly advanced, significantly improving patients' access to reliable health-related information. However, current approaches face notable challenges, including the difficulty in obtaining large-scale and unbiased medical datasets, significant privacy concerns, and inefficiencies due to manual dataset annotation. To address these issues, This study introduce a novel methodology leveraging publicly accessible online health forums to systematically build an unbiased, privacy-conscious QA dataset, and it employs Topic-guided Semantic Modeling (TGSM) for automated topic identification, enabling efficient and targeted annotation of relevant patient-generated content. Subsequently, this study propose a two-stage QA pipeline based on a Retriever–Reader architecture, which is further enhanced through fine-tuning state-of-the-art transformer-based models on the constructed domain-specific QA dataset. Experimental results demonstrate that our fine-tuned BioBERT significantly outperforms existing benchmarks, offering accurate patient-derived insights and providing a replicable framework for building efficient medical QA systems.

\section*{Author summary}

In recent years, question-answering (QA) systems leveraging natural language processing (NLP) have significantly improved how patients access reliable health information. However, developing effective QA systems in healthcare remains challenging due to difficulties in obtaining large-scale unbiased datasets, privacy concerns, and the labor-intensive nature of manual dataset annotation. In this paper, we propose a method to systematically build an unbiased and privacy-conscious QA dataset by leveraging publicly accessible online health forums, where individuals openly share their medical experiences. To enhance dataset construction, this work applies a Topic-guided Semantic Modeling (TGSM) technique, efficiently identifying topics and corresponding aspects from unstructured patient-generated content, thus simplifying and accelerating the annotation process. This study then implements a two-stage QA pipeline featuring a Retriever-Reader architecture and further optimizes it by fine-tuning advanced transformer-based models on our dataset. Our experimental results demonstrate that our fine-tuned BioBERT model significantly outperforms existing benchmarks, highlighting the effectiveness and robustness of our approach. By providing accurate patient-derived insights and a replicable framework, our method can contribute meaningfully to the development of effective QA systems for medical applications.


\section{Introduction}
\label{sec:intro}

The widespread adoption of the internet has enabled the growth of diverse online communities in which individuals exchange information, seek support, and share personal experiences \cite{nimrod2013prosuming,pendry2018online}. Among these, online health and illness-related forums have become particularly important for individuals living with chronic or stigmatized conditions, as they provide spaces for asking questions, discussing symptoms, and receiving informational and emotional support from others with similar experiences \cite{white2001receiving,moorhead2013new,marshall2024forums,marshall2025safety}. In the context of mental health, such forums may reduce isolation, normalize lived experiences, and facilitate access to peer-generated advice; however, their value also depends on issues such as psychological safety, moderation, and the quality of the information exchanged \cite{marshall2024forums,hao2019providing}. These characteristics make online mental-health communities both socially important and methodologically valuable for computational health research. Beyond their supportive function, patient forums constitute a large repository of first-hand, patient-generated narratives that can inform the development of data-driven tools for health information access and decision support \cite{van2014effectiveness}. Compared with many clinical data sources, publicly available forum content may offer broader coverage of everyday concerns, symptom descriptions, treatment experiences, and practical coping strategies expressed in natural language. This is particularly relevant for schizophrenia, a complex psychiatric disorder whose lived experience often includes symptoms, medication-related concerns, stigma, functional impairment, and recurrent uncertainty in everyday life. Recent NLP studies have further shown that psychosis-related online text contains measurable linguistic patterns that can be analyzed computationally, underscoring the value of publicly available mental-health discourse for scalable language-based research \cite{plank2024coherence}. Nevertheless, although these data are increasingly studied for classification and linguistic analysis, they remain underutilized for the development of domain-specific question-answering resources grounded in patient experience. Against this backdrop, question answering (QA) has emerged as a particularly important natural language processing (NLP) paradigm for improving access to health information by enabling users to obtain relevant answers to their information needs from large and diverse textual sources \cite{lu2021natural,gonzalez2021natural}. In general, QA systems aim to answer natural-language questions by identifying, extracting, ranking, or generating relevant answers from one or more textual sources \cite{khashabi2020unified}. Their success depends not only on the model architecture, but also on the quality of the underlying data, the realism of the questions, and the alignment between the training objective and the target use case \cite{hao2022recent,kell2024qa}. The introduction of transformer-based language models, especially BERT and its biomedical variants, substantially improved machine reading comprehension and biomedical QA by enabling a two-stage paradigm of large-scale pre-training followed by task-specific fine-tuning \cite{c2,c4,cho2024transformer}. More recently, large language models and retrieval-augmented approaches have pushed medical QA even further, demonstrating strong performance on benchmark datasets and increasingly complex reasoning tasks \cite{singhal2025expert}. However, these advances do not eliminate the need for carefully constructed, domain-specific datasets, particularly when the target setting involves specialized disorders, consumer-oriented language, and real-world patient concerns.

Recent evidence suggests that important gaps remain between progress on benchmark medical QA and the development of systems that are genuinely useful in practice. A recent systematic review of QA systems for health professionals found that many systems rely on restricted or unrealistic question sources, with only a limited number of studies using genuine real-world questions, and that most systems still fall short in terms of transparency, utility, and clinical applicability \cite{kell2024qa}. Related research on consumer health question quality has shown that questions posted in web-based health communities vary substantially in clarity, completeness, and answerability, which directly affects the usefulness of retrieved or generated responses \cite{alasmari2024quality}. These limitations are even more pronounced in mental-health settings, where questions are often subjective, context-dependent, emotionally charged, and potentially associated with multiple valid answers. As a result, despite the growth of biomedical QA, there remains a lack of disorder-specific QA resources that are derived from authentic patient discourse and designed to capture the everyday informational needs of people living with schizophrenia. Constructing such a resource is challenging for several reasons. First, obtaining sufficiently rich and representative mental-health text corpora is difficult because of privacy concerns, restricted access to clinical data, and the sensitive nature of psychiatric information \cite{krishna2021hurdles}. Second, annotation of QA datasets in specialized medical domains is labor-intensive and requires decisions about topic boundaries, question formulation, answer span selection, and handling of instances in which multiple answers may be acceptable \cite{krishna2021hurdles,kell2024qa}. Third, pre-processing and organizing patient-generated text is itself nontrivial, because forum posts are often informal, heterogeneous in structure, and distributed across many latent themes. For schizophrenia-related data, in particular, this complexity is amplified by the breadth of topics discussed, ranging from hallucinations and delusions to medication effects, social functioning, work, education, and relationships.

To address these gaps, this study develops a schizophrenia-specific QA framework grounded in publicly available patient-forum data. Rather than relying on highly curated clinical notes or examination-style questions, this study focuses on naturally occurring discussions authored by individuals affected by schizophrenia and related concerns. First, this study introduce an effective approach for collecting data on a particular medical condition to build a comprehensive dataset while avoiding privacy concerns. By scraping a web forum dedicated to schizophrenia patients, it is feasible to obtain relevant information about daily problems and symptoms experienced by them, such as which actions or activities are most commonly affected by the disease, common hallucinations, etc. Since forums are typically frequented by people from diverse places and different backgrounds and experiences, the sample is extensive and significant. Another advantage is the openness of such forums, where people feel free to express their thoughts and experiences without judgment, enabling the capture of a more in-depth, comprehensive array of patients' experiences living with the disease. Second, this study optimize and accelerate the QA annotation process by leveraging a specialized annotation tool designed to efficiently support NLP-based QA and information retrieval tasks. Then, Topic-guided Semantic Modeling (TGSM) is employed to identify topics within the corpus automatically. Questions are identified in the QA dataset based on these topics, with questions possibly having multiple answers corresponding to the most relevant aspects contained within that topic. Finally, this study shows how to optimally set up a pipeline for using a QA model and how to significantly improve the model's efficiency by using the Retriever \cite{c7} filter. This filter can quickly examine an entire archive of documents and pass the most relevant documents as the response. Through experiments with the BioBERT \cite{c39}, RoBERTa \cite{c14} and DistilBERT \cite{c29} models in our pipeline, this study empirically demonstrate the boost in performance of these models in schizophrenia-related tasks by fine-tuning with the schizophrenia QA dataset, reaching the state-of-the-art in the field of QA models for mental health issues. The main contributions of this paper are the following:

\begin{itemize}
   \item A novel method of obtaining data on a specific medical disease, such as schizophrenia, to construct an unbiased, easy-to-obtain dataset without privacy issues is presented. By web-scraping dedicated patient forums frequented by people from diverse locations, backgrounds, and experiences, the constructed dataset offers a diverse, comprehensive, and significant sample size while ensuring data authenticity. 

    \item This study optimize and accelerate the question-answering annotation process. By processing the corpus of user posts  to first identify topics, and then selecting posts in the corpus most relevant to each topic to serve as a basis for annotating questions and answers for the QA dataset, we can significantly speed up this process.

    \item This study illustrate establishing an efficient QA pipeline to utilize a Retreiver-Reader model and present results with the BioBERT, RoBERTa, and DistillBERT models. Extensive empirical results show the potential to enhance these models' performance further within the domain of schizophrenia as a result of fine-tuning with the schizophrenia QA dataset, leading to state-of-the-art results for mental health QA models.
\end{itemize}

\section{Background}
\label{sec:related_work}

The work on data mining techniques for application in medical fields is closely related to ours, particularly techniques for obtaining large medical datasets, constructing QA datasets, and implementing QA models. We discuss each in detail below.

\subsection{Data for medical applications}
\label{subsec:medical_data}
One common method for obtaining medical data involves extracting information from clinical charts, but such data are difficult to access due to privacy concerns \cite{c6}. \cite{c8} used telemedicine dialogs between doctors and patients. However, this approach requires extensive pre-processing, and collecting enough such data in a privacy-preserving manner is also challenging. In \cite{c9}, the authors collect data from professional medical board exams. While a valuable source of information, it lacks information shared directly by patients, thus lacking the depth and nuanced knowledge to be gained from patients' lived experiences of a disease. \cite{c32} presents one of the most interesting methods proposed so far, using QA forums to create a QA dataset utilizing selected questions in the medical/biological field on the Reddit web platform and using the answers with the highest score. Nevertheless, the problem of a lack of information directly from patients affected by the disease under consideration persists. In addition, this method is not well-suited to allow multiple answers to a question. To improve on these limitations of current methods, we present a novel method to construct a comprehensive, nuanced QA dataset that is easy to obtain, has low noise, and contains patients information by scraping conversations from schizophrenic patients on ``The Mental Health Forum''.

\subsection{Development of a QA dataset}
\label{subsec:qa_dataset_dev}

A previous work \cite{c33} constructed a QA dataset for COVID-19 and used it to train the COVID-QA model. However, it uses very long text documents and only supports one answer to a question. Other recent work demonstrate similar limitations, for instance, the RxWhyQA dataset was developed for clinical QA and designed to handle multi‑answer and multi‑focus questions, but it still comes from structured clinical notes rather than patient‑generated posts \cite{bardhan2024question}. Another work by adapting pre‑trained generative Models for extractive question answering, leveraged generative models for extractive QA but remains oriented to generic corpora rather than disease‑specific forums \cite{mallick2023adapting}. In contrast, our method takes advantage of Latent Dirichlet Allocation's ability to give good results in identifying topics contained in text files \cite{c15,c16,c17,c18}. By first applying TGSM to systematically discover relevant topics and aspects within our corpus, this study effectively streamlines the annotation process. Instead of exhaustively annotating all possible questions in unstructured text, our annotators focus exclusively on generating and labeling questions and answers aligned with previously identified topics. This targeted approach efficiently addresses limitations observed in earlier methods \cite{c33}, as our resulting dataset naturally accommodates multiple concise and contextually relevant answers per question, each derived from individual patient posts that typically span only a few short paragraphs.

\subsection{Advances in QA model for medicine}
\label{subsec:medical_qa_model}

Pre-trained BERT models can be fine-tuned to create a QA model tailored to a specific domain \cite{c11,c2,c13,c14}. Numerous domain-specific QA models have been developed, with previous work notably involving fine-tuning BERT variants for schizophrenia-related applications \cite{c36,c37}. Recently, significant progress has been made with transformer-based models such as DistilBERT, RoBERTa and BioBERT \cite{c39}, demonstrating effectiveness in biomedical question-answering tasks. Specifically, BioBERT \cite{c39}, a benchmark in medical QA, has been pre-trained on extensive biomedical literature and clinical data, yet remains broadly focused rather than disease-specific. In parallel, substantial advancements have occurred in large language models (LLMs) such as GPT, Med-PaLM, and LLaMA, enabling deeper contextual understanding and accurate and coherent medical response generation from minimal prompts \cite{c6}. Med-PaLM~\cite{tu2024towards}, an adaptation of Google's Pathways Language Model (PaLM), achieves near-expert performance in answering intricate clinical questions, significantly bridging the gaps between AI-generated and human clinical expertise. Similarly, GPT and LLaMA have also demonstrated proficiency in medical QA tasks, improving healthcare interactions with nuanced and contextually aware responses. Despite these notable advances, the practical implementation of LLMs in medical QA faces ongoing challenges, primarily due to dataset limitations and validation issues. The creation of reliable, clinically validated, and unbiased datasets continues to be a critical bottleneck, specifically traditional data sources such as clinical charts or medical examinations lack detailed patient-generated content, while online patient forums, although rich in firsthand experiences, present challenges regarding data reliability, privacy, and ethical concerns \cite{c14,c36}.

\section{Materials and Methods}
\label{sec:method}

In this section, this study first presents the methods used for collecting and pre-processing data related to schizophrenia and how these data were used to construct a schizophrenia QA dataset. Next, this study describes the pipeline structure of our question-answering system, as well as how it was used and optimized. This study then illustrates the fine-tuning process of BERT models for use in applications related to schizophrenia. Afterward, it describes the experimental setup and the metrics used and concludes the section with the experimental results of three different models (DistilBERT \cite{c29}, RoBERTa-base-squad2 \cite{c14}, and BioBERT \cite{c39}) and their versions fine-tuned for schizophrenia.

\subsection{Acquisition and pre-processing of medical data}
\label{subsec:medical_data}

Creating a domain-specific QA dataset for mental health requires a reliable source of natural language data reflecting patients' experiences. Web health forums dedicated to mental diseases can provide such content in abundance. One such forum is the Mental Health Forum \footnote{\url{https://www.mentalhealthforum.net/}}, which has a subforum dedicated to schizophrenia, where patients with schizophrenia, as well as their families and friends, periodically share their personal experiences with the disease. The original corpus consists of 415,602 posts, each labeled with its post ID, date of posting, and poster. Identifying information (such as the poster username) was deleted from the data.

In this study, the forum-derived corpus is considered unbiased in a comparative methodological sense. Unlike data collected through single-site clinical recruitment, interviewer-mediated interactions, or predefined questionnaires, forum posts represent naturally occurring narratives that are not shaped by clinician prompting, survey design, or other researcher-imposed elicitation constraints. This allows the corpus to capture a broader range of spontaneously expressed symptoms, daily difficulties, medication-related concerns, and lived experiences. At the same time, participation in online forums is inherently self-selected, and the resulting corpus may therefore still reflect sampling bias associated with internet access, engagement in peer-support communities, and willingness to disclose personal experiences publicly. Accordingly, the term ``unbiased'' is used here to indicate reduced institutional and elicitation bias rather than the complete absence of all forms of sampling bias. The proposed data-collection strategy is also privacy-conscious. In addition to removing usernames, the preprocessing pipeline excluded direct personal identifiers when present, retained only the textual content necessary for topic modeling, QA construction, and downstream evaluation, and avoided preserving unnecessary user-level metadata. No attempt was made to infer, reconstruct, or link real-world identities from the collected posts. Because the corpus was derived exclusively from publicly accessible forum content and processed in de-identified textual form, these measures help reduce privacy risks while still enabling the use of patient-centered online discourse for domain-specific QA development.

Several text pre-processing steps were implemented to prepare the data for subsequent topic analysis. Initially, unnecessary words and characters were removed to make the aspect words meaningful for clustering. Stop words, punctuation, and numbers were also eliminated. However, while most stop words were removed, negation words (\textit{e.g.,} ``no'' or ``not'') were specifically retained. This was crucial because the negation words significantly impact the semantic meaning of sentences and thus influence the results of subsequent topic modeling. Following this, we perform stemming of words, making them lowercase, and  branching of words, resulting in a shortening at the root. For example, ``apple'' and ``apples'' both become ``appl'' and are treated as the same word in the vectorization step.

After completing these initial pre-processing steps, we segmented the cleaned forum corpus into paragraph-level textual units and organized them according to their thematic content. Since the original forum data were large-scale, informal, and highly heterogeneous, a topic-level organization step was necessary before manual QA annotation. To this end, we employed Topic-guided Semantic Modeling (TGSM), an unsupervised topic modeling strategy designed to discover latent semantic structures from domain-specific patient-generated text. Similar to Latent Dirichlet Allocation (LDA)~\cite{xu2023short}, TGSM assumes that each textual unit can be represented as a probabilistic mixture of latent topics and that each topic can be characterized by a probability distribution over words. However, the role of TGSM in our framework extends beyond standard corpus-level topic discovery. Rather than using topic modeling only to describe the overall corpus, TGSM uses the inferred topic distributions and high-probability aspect terms to organize the corpus into topic-centered paragraph clusters, which subsequently guide question formulation and answer-span annotation. In this sense, TGSM serves as both a topic discovery method and a semantic filtering mechanism for constructing the QA dataset.

This topic-guided design also explains why TGSM is more suitable for our annotation workflow than directly applying standard LDA or purely embedding-based clustering. Compared with standard LDA, TGSM is integrated more explicitly with the downstream QA construction process: the discovered topics are not only interpreted after modeling, but are also used to structure the annotation space and identify aspect-level cues for question generation. Compared with embedding-only clustering, TGSM provides more transparent topic-aspect associations, since each topic is represented through explicit high-probability words or phrases. This interpretability is particularly important in a sensitive medical domain, where annotators need to understand why a group of paragraphs is assigned to a particular theme before formulating questions and selecting answer spans. Therefore, TGSM allows the annotation process to remain grounded in patient-generated language while reducing the need to manually screen the entire corpus.

Based on this topic-guided framework, we further determined the appropriate number of topics through a combined quantitative and qualitative procedure. We first trained TGSM models using multiple candidate topic numbers and evaluated their goodness-of-fit using held-out perplexity. However, perplexity alone may not fully capture whether the resulting topics are meaningful for human annotation. Therefore, we also examined topic coherence by checking whether the top-ranked aspect terms within each topic were semantically consistent and whether representative paragraphs assigned to the same topic reflected a coherent schizophrenia-related theme. Models with fewer topics tended to merge distinct concerns, such as hallucinations, medication effects, and social functioning, into overly broad categories. In contrast, models with substantially more topics produced fragmented or redundant themes that were less useful for annotation~\cite{c40}.

\begin{table*}[h]
\centering
\caption{Example of topics and aspects.}
\vspace{+0.2cm}
\label{tab:topic}
\begin{tabular}{|l|l|}
\hline
 \textbf{Topics} &  \textbf{Aspects} \\
\hline
1 &  ``hallucinations'', ``afraid of'', ``memory'', ``problems'', ...\\
2 &  ``stop'', ``caffè'', ``nicotine'', ``drinking'', ... \\
3 &  ``pandemic'', ``suffering'', ``society'', ``family'', ...\\
4 &  ``food'', ``weight'', ``struggle'', ``being clean'', ...\\
...& ...\\
\hline
\end{tabular}
\end{table*}

The topic analysis performed using TGSM resulted in 35 distinct topics and 300 relevant aspects. Specifically, the optimal number of topics was determined by evaluating the goodness-of-fit, measured by perplexity, across multiple candidate topic counts on a held-out validation set. Among these candidate values, 35 provided the best balance between model interpretability and minimization of perplexity. Subsequently, for each identified topic, the top-ranked keywords or phrases, based on their probabilistic association with the respective topic, were selected as ``aspects''. By aggregating these top keywords across all 35 topics, we obtained a comprehensive set of approximately 300 unique aspects, each representing a distinct sub-theme or detailed semantic element within the broader thematic categories. An illustration of the results of the topic analysis, including examples of topics and their corresponding aspects, along with their associations with schizophrenia-related problems and symptoms, is presented in Table~\ref{tab:topic}. Furthermore, the corpus posts were then segmented into paragraphs and organized by these identified topics, facilitating systematic and efficient annotation of the corresponding questions and answers by the researchers. This structured approach also enhanced retrieval efficiency and QA performance.

\subsection{Development of a domain-specific QA dataset}
\label{subsec:domain_qa}

\begin{figure*}[h]
    \centering
    \includegraphics[width=0.85\linewidth]{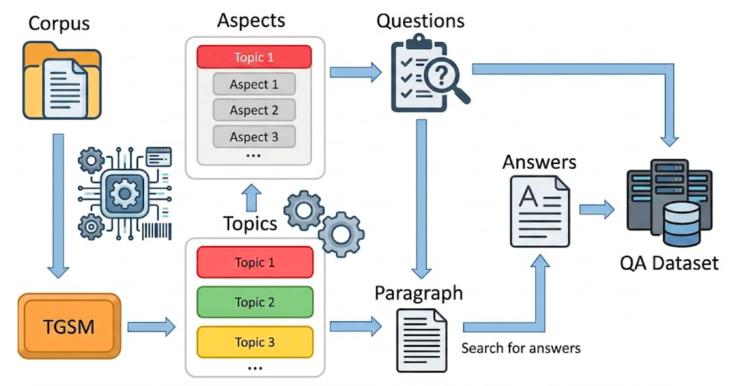} 
    \caption{Overview of the workflow for constructing QA datasets. The corpus is initially analyzed using TGSM topic modeling to identify semantic topics and their associated aspects. These identified aspects guide the creation of targeted questions, which are subsequently matched with corresponding answers extracted from paragraphs organized by topic, ultimately forming structured QA datasets.}
    \label{fig:my_label}
\end{figure*}

After TGSM identified 35 topics and their associated aspects, the most relevant characteristics within each topic were examined in the corresponding topic-organized paragraphs. Based on these aspect-topic associations, natural language questions were formulated, and answer spans were manually annotated from the relevant passages. For instance, the aspect ``afraid of'' yielded questions such as ``What is a schizophrenic afraid of?'', whereas the aspect ``drinking'' yielded context-specific questions such as ``What does a schizophrenic stop with?''.  The outline of our method is shown in Fig~\ref{fig:my_label}.


Within each topic-centered paragraph cluster, annotators reviewed the representative paragraphs and the corresponding TGSM-derived aspect terms to formulate natural-language questions and identify answer spans. The annotation standard required each question to be grounded in a coherent topic-aspect association and each answer to be explicitly supported by the corresponding paragraph evidence. Because patient-generated forum discussions often contain multiple valid expressions of similar experiences, multiple answers were retained when they were textually supported and semantically relevant. For each of the 35 topic clusters, 30 question-answer pairs were constructed, resulting in a total of 1,050 QA pairs. This process explains how the large-scale corpus of 415,602 posts was reduced into a compact, topic-guided QA dataset: the full corpus provided the source material, TGSM organized the corpus into 35 semantically coherent topic clusters, and manual annotation produced 1,050 QA pairs from the representative paragraph evidence within those clusters.

Consequently, the dataset construction process should be understood as a topic-guided sampling and annotation procedure rather than a direct compression of all posts into a small number of paragraphs, as shown in Table~\ref{tab:dataset_stat}. The purpose of TGSM was to reduce the manual search burden by narrowing the annotation space from the full forum archive to semantically organized topic clusters. This design allowed the annotators to focus on clinically and experientially meaningful schizophrenia-related themes while preserving the ability to annotate multiple concise answers for a single question.

\begin{table}[h]
\centering
\caption{Statistics of the schizophrenia QA dataset.}
\vspace{+0.2cm}
\begin{tabular}{|l|l|l|}
\hline
\textbf{Dataset}& \textbf{Element} & \textbf{Count} \\
\hline
    \multirow{4}{*}{Schizophrenia QA}&User posts & 415,602  \\
    &Max seq words & 385 \\
    &Topics paragraph & 35 \\
    &Question and Answer & 1,050\\
\hline
\end{tabular}
\label{tab:dataset_stat}
\end{table}

\subsection{QA pipeline}
\label{subsec:qa_pipeline}



The corpus is divided into paragraphs based on identified topic clusters to provide structured inputs for the QA pipeline. Several text-cleaning operations were applied to ensure high-quality input data, including normalization of consecutive blank lines, removal of leading and trailing whitespace from text lines, elimination of repetitive headers and footers, and careful adjustment of overlaps between adjacent paragraphs to preserve sentence integrity during document segmentation.

Following pre-processing, we implemented a topic-guided annotation strategy to construct the domain-specific QA dataset. We utilized topic modeling TGSM to cluster the corpus into 35 distinct topics based on semantic patterns. Human annotators then examined these topic clusters to identify key ``aspect'' (\textit{e.g.,} symptoms, side effects) and formulated natural language questions corresponding to these aspects (\textit{e.g.,} ``What is a schizophrenic afraid of?''). Relevant answer spans were then manually annotated within the constituent paragraphs. This semi-automated approach facilitated the efficient creation of 1,050 question-answer pairs, ensuring that the dataset captures the nuanced, multi-answer nature of patient experiences.

To effectively utilize this annotated corpus and promote accurate and efficient QA processing, our pipeline primarily consists of two components: a Retriever and a Reader. 

\begin{itemize} 

\item The \textbf{Retriever} serves as a lightweight and fast filtering mechanism. It scans the entire document archive to identify a subset of candidate documents that are most relevant to the user's query. By leveraging the pre-computed topic clusters, the Retriever sifts through irrelevant documents, significantly reducing the computational overhead for the subsequent stage. In our implementation, the Retriever is configured to filter the corpus based on the topic-based paragraphs, ensuring that the search space is narrowed down efficiently before detailed processing.

\item The \textbf{Reader} performs the core machine comprehension task. We employed the \texttt{FARMReader} component from the Haystack library, which is designed to handle transformer-based models. The Reader takes the candidate documents provided by the Retriever and performs a deep syntactic and semantic analysis to extract the precise text span that best answers the query. For this study, the Reader is powered by fine-tuned instances of pre-trained models, including BioBERT, RoBERTa, and DistilBERT, allowing it to adapt to the specific biomedical and colloquial context of the schizophrenia dataset, ensuring high accuracy and relevance in the extracted answers.

\end{itemize}

The pipeline performance depends significantly on two parameters: the Retriever top-$k$ and Reader top-$k$. Specifically, the Retriever top-$k$ determines how many candidate documents are passed to the Reader, reflecting a trade-off between retrieval coverage and computational efficiency. A higher top-$k$ ensures fewer missed relevant documents but increases computational demand. Meanwhile, the Reader top-$k$ parameter defines how many of the top-ranked answer candidates are finally presented. Through empirical experiments, we identified the optimal parameter settings as Retriever top-$k=35$, aligning with the number of TGSM topics, and Reader top-$k=10$, providing the best balance between retrieval recall and QA accuracy.

\section{Expreiment}
\label{sec:result}
In this section, this work first describes the experimental setup, including model configurations, evaluation metrics, and datasets, then presents and discusses the experimental results obtained from evaluating our proposed QA pipeline.

\subsection{Experimental Setup}
\label{subsec:exp_setup}

The BERT (Bidirectional Encoder Representations from Transformers) models were chosen for their ability to be easily adapted to the specific domain of interest \cite{c6,c7,c9}. The models DistilBERT, RoBERTa, and BioBERT were fine-tuned and compared with their pre-trained versions. 

\begin{itemize}
    \item \textbf{DistilBERT} \cite{c29} is a transformer model, smaller and faster than BERT, which was pre-trained on the same corpus in a self-supervised fashion, using the BERT base model as a teacher. This means it was pre-trained on the raw texts only, with no humans labeling them in any way, with an automatic process to generate inputs and labels from those texts using the BERT base model. DistilBERT was pre-trained on BookCorpus, a dataset of 11,038 unpublished books, and English Wikipedia, the same data as BERT.
    \item \textbf{RoBERTa} \cite{c14} is a transformer model pre-trained on a large corpus of English data in a self-supervised manner. In other words, it was trained on the raw texts only, with no humans labeling them in any way, with an automatic process to generate inputs and labels from those texts. The RoBERTa model was pre-trained on the union of five datasets: BookCorpus, a dataset consisting of 11,038 unpublished books; English Wikipedia excluding lists, tables, and headers; CC-News, a dataset containing 63 million English news articles crawled between September 2016 and February 2019; OpenWebText, an open-source recreation of the WebText dataset; and Stories, a dataset containing a subset of CommonCraw data filtered to match the story-like style of Winograd schemas. Together, these datasets constitute 160GB of text.
    \item \textbf{BioBERT} \cite{c39} is a domain-specific language representation model pre-trained on large biomedical corpora. BioBERT vastly outperforms BERT and previous state-of-the-art models in various biomedical text mining tasks when pre-trained on biomedical corpora. In our experiments, this model is used as a benchmark for our comparison, since it is a model trained on medical data that achieves results in medical tasks equal to the current state of the art. The model was trained on all MIMIC notes for 150,000 steps.
\end{itemize}

For domain-specific datasets, domain adaptation, \textit{i.e.}, tuning the pre-trained model on customized examples, has been proven to increase performance even with relatively small datasets. Depending on the domain, tuning on 2,000 examples can lead to a performance boost of roughly 5\% to 20\% \cite{c7}. We trained the fine-tuned QA model using the Transformers library from Hugging Face \cite{c42} and the FARMReader method from the Haystack library \cite{c7}. To achieve domain adaptation, the models are trained on the Schizophrenia QA dataset. The models were trained on 900 questions and answers annotated on the specific-domain QA dataset for schizophrenia. The questions represented the entirety of the 35 topics previously identified, utilizing 70\% of the QA dataset. Table~\ref{tab:hyperparams} shows the settings and hyperparameters used for training the model.

\begin{table}[h]
\centering
\caption{Experimental settings and hyperparameters used for training the model.}
\vspace{+0.2cm}
\label{tab:hyperparams}
\begin{tabular}{|l|l|}
\hline
\textbf{Parameter} &  \textbf{Value} \\
\hline
        Optimizer   &  ADAM\\
        Weight decay    &  0.01  \\
        Learning rate   &  1e-05     \\
        Num of train epochs & 20 \\
        Max seq length  & 384 \\
        Max length  after clipping  & 256 \\
        Average length after clipping   &   255.90 \\
        Examples in train & 386,001  \\ 
\hline
\end{tabular}
\end{table}

\subsection{Evaluation metrics}
\label{subsec:eval_metric}

The performance of the proposed models was evaluated using four standard metrics: Precision (Equation~\ref{eq:prec}), Recall (Equation~\ref{eq:rec}), F1-score (Equation~\ref{eq:f1}) and Exact Match (EM). The evaluation was conducted on a held-out test set, comprising 30\% of the total QA dataset.

\noindent \textbf{Precision} measures the proportion of correctly predicted answers among all answers predicted by the model. Ranges from 0 to 1, where a higher value indicates that the model's predictions are more reliable. Formally, precision is defined as:

\begin{equation}
\label{eq:prec}
\textrm{Precision} = \frac{\textrm{TP}}{\textrm{TP}+\textrm{FP}}
\end{equation}

\noindent where TP (True Positives) denotes the number of answers correctly identified as relevant, and FP (False Positives) represents the number of answers incorrectly identified as relevant.

\noindent \textbf{Recall} evaluates the Retriever component's capability to include documents containing correct answers within the retrieved candidates. Specifically, recall measures the fraction of correctly identified relevant documents out of all documents that actually contain correct answers. Recall also ranges from 0 to 1, with 1 indicating perfect retrieval performance:

\begin{equation}
\label{eq:rec}
\textrm{Recall} = \frac{\textrm{TP}}{\textrm{TP}+\textrm{FN}}
\end{equation}

\noindent where FN (False Negatives) represents the number of relevant documents the retriever failed to retrieve.

\noindent \textbf{F1-score} combines precision and recall into a single metric through their harmonic mean, thus balancing both types of errors (false positives and false negatives). It is particularly suitable when aiming to achieve an effective balance between precision and recall. The F1-score is defined as follows:

\begin{equation}
\label{eq:f1}
\textrm{F1-score} = 2\times\frac{\textrm{Precision}\times\textrm{Recall}}{\textrm{Precision}+\textrm{Recall}}
\end{equation}

\noindent An F1-score value close to 1 indicates high precision and recall, while a value close to 0 suggests poor performance in both metrics.

\noindent \textbf{Exact Match (EM)} calculates the percentage of predicted answers that exactly match the ground truth answers. It assigns a binary outcome to each prediction: 1 for an exact textual match and 0 otherwise. For instance, given the ground truth answer ``He is afraid to leave the house'' to the question ``What is a schizophrenic afraid of?'', a predicted answer such as ``He is afraid of leaving the house'' does not yield an EM score of 1, since it is not identical to the ground truth.

\subsection{Evaluation datasets}

To systematically evaluate the generality and effectiveness of our proposed QA pipeline, we employ three datasets: our schizophrenia-specific QA dataset, the MHQA mental health benchmark, and the PubMedQA biomedical research benchmark.

\noindent \textbf{Schizophrenia QA dataset:} This dataset was constructed using patient-generated content from ``The Mental Health Forum'', specifically targeting schizophrenia-related discussions. It contains 415,602 user posts segmented into paragraphs grouped by topic using TGSM. Questions and multiple short answers were annotated based on identified aspects within these topics, capturing nuanced patient experiences related to symptoms and daily impacts of schizophrenia. Additional dataset details and annotation procedures are described in previous sections.

\noindent \textbf{MHQA \cite{joshi2025mhqa}:} MHQA is a recent knowledge-intensive mental health QA benchmark designed to evaluate language models on their ability to handle diverse, expert-level mental health questions. It contains approximately 58,600 multiple-choice questions spanning four major mental health conditions: anxiety, depression, trauma, and obsessive-compulsive issues. Each question is accompanied by evidence derived from PubMed abstracts. MHQA provides both an expert-annotated subset (MHQA-Gold, 2,475 questions) and a larger pseudo-labeled set (MHQA-B). In our evaluation, we specifically utilize MHQA-Gold to ensure high-quality benchmarks and reliable comparisons.

\noindent \textbf{PubMedQA~\cite{jin2019pubmedqa}:} PubMedQA is a widely used biomedical QA dataset developed from research articles available in PubMed. It consists of research questions generated from paper titles, paired with corresponding abstracts as context. Each instance requires models to determine a final answer from three categories: ``yes'', ``no'' or ``maybe'' reflecting conclusions drawn from the abstracts. PubMedQA comprises a carefully annotated subset PQA-L (1,000 QA pairs), along with larger unlabeled and automatically labeled subsets. For our experiments, we focus on the expert-annotated subset to rigorously evaluate the performance of our QA pipeline in general biomedical contexts beyond mental health forums.

\subsection{Experimental results}
\label{subsec:result}

Extensive experiments were conducted by fine-tuning three different pre-trained models: DistilBERT, RoBERTa, and BioBERT, across three medical QA datasets: MHQA, PubMedQA, and Schizophrenia QA. The comprehensive results are presented in Table~\ref{tab:result}. Consistently, the Fine-tuned-BioBERT model achieved the best performance across all datasets and metrics. In the case of the MHQA and PubMedQA datasets, the fine-tuning process yielded substantial gains. For MHQA, the Fine-tuned-BioBERT achieved a Precision score of 0.781 and an F1-score of 0.772, representing a notable improvement over the base BioBERT model. Similarly, on the PubMedQA dataset, the fine-tuned BioBERT reached a top Precision of 0.816. It is worth noting that even the base BioBERT model outperformed the fine-tuned versions of DistilBERT in certain metrics, highlighting the importance of domain-specific pre-training in biomedical tasks. The most significant performance boost was observed in the Schizophrenia QA dataset. There is a 14.30\% increase in the Precision score between the base BioBERT model and its fine-tuned version, and a remarkable 34.4\% increase for the Exact Match (EM) score, jumping from 0.427 to 0.617. When comparing the lightest model, DistilBERT, with the best-performing Fine-tuned BioBERT on this dataset, there is a boost of 32.59\% in terms of Precision score. This trend is consistent across other metrics, with Fine-tuned BioBERT showing a superior ability to retrieve accurate answers. The results clearly demonstrate that while model architecture plays a role, fine-tuning on specific medical corpora is the decisive factor for maximizing Precision and EM scores.

\begin{table*}[h]
\centering
\caption{Evaluation results of DistilBERT, RoBERTa, and BioBERT models across MHQA, PubMedQA, and our Schizophrenia QA dataset. Best results per dataset are in bold.}
\vspace{+0.2cm}
\label{tab:result}
\begin{adjustbox}{max width=\textwidth}
\begin{tabular}{|l|l|l|l|l|l|}
\hline
\textbf{Dataset} & \textbf{Models} & \textbf{Precision (\%)} & \textbf{F1-score (\%)} & \textbf{Recall (\%)} & \textbf{EM (\%)} \\
\hline
\hline
\multirow{6}{*}{MHQA}
& DistilBERT            & 58.2 & 54.1 & 60.3 & 30.2 \\
& RoBERTa               & 62.4 & 58.7 & 65.3 & 33.1 \\
& BioBERT               & 66.3 & 61.8 & 68.1 & 35.9 \\
& Fine-tuned DistilBERT & 70.9 & 67.6 & 73.2 & 40.3 \\
& Fine-tuned RoBERTa    & 75.6 & 72.8 & 78.1 & 45.1 \\
& Fine-tuned BioBERT    & \textbf{78.1} & \textbf{76.2} & \textbf{82.7} & \textbf{49.7} \\
\hline
\hline
\multirow{6}{*}{PubMedQA}
& DistilBERT            & 60.1 & 56.3 & 61.7 & 35.2 \\
& RoBERTa               & 64.2 & 59.2 & 65.9 & 37.7 \\
& BioBERT               & 67.5 & 62.1 & 68.8 & 39.5 \\
& Fine-tuned DistilBERT & 73.2 & 69.1 & 74.7 & 43.8 \\
& Fine-tuned RoBERTa    & 77.4 & 73.5 & 79.1 & 47.2 \\
& Fine-tuned BioBERT    & \textbf{81.6} & \textbf{77.9} & \textbf{83.4} & \textbf{51.2} \\
\hline
\hline
\multirow{6}{*}{Schizophrenia QA}
& DistilBERT            & 68.1 & 63.9 & 71.3 & 42.5 \\
& RoBERTa               & 74.6 & 74.3 & 77.2 & 42.5 \\
& BioBERT               & 79.0 & 77.5 & 88.0 & 42.7 \\
& Fine-tuned DistilBERT & 80.5 & 78.3 & 83.1 & 43.8 \\
& Fine-tuned RoBERTa    & 88.5 & 80.3 & 86.1 & 50.8 \\
& Fine-tuned BioBERT    & \textbf{90.3} & \textbf{88.5} & \textbf{91.6} & \textbf{57.7} \\
\hline
\end{tabular}
\end{adjustbox}
\end{table*}

\noindent Table~\ref{tab:qa_example} contains examples of interactions with the fine-tuned BioBERT model. The questions asked concern the problems and symptoms. Evaluating the responses provided by the model allows us to conclude that the reported responses are relevant and logical. It is also clear that the model can provide multiple relevant answers for each question.

\begin{table*}[hbt!]
\caption{Question-Answer examples obtained with the fine-tuned BioBERT model.}
\vspace{+0.2cm}
\label{tab:qa_example}
\begin{adjustbox}{max width=\textwidth}
\begin{tabular}{|l|l|}
\hline
   \textbf{Question} & \textbf{Answer} \\
\hline
  \multirow{2}{*}{What is a schizophrenic afraid of?}& afraid of sanity / afraid of outside / afraid\\ & of hallucinations \\ 
  What is a schizophrenic obsessed with? & obsessed with food / conspiracy theories/ rabies \\
  What is a schizophrenic suffering from? & being tired all the time / insomnia / delusions \\
 \multirow{2}{*}{What does a schizophrenic stop with?}& smoking cigarettes / stop vomiting / stop drinking \\ &  and drugs \\  
  What does a schizophrenic struggle with? & hallucinations / medication / people \\
  What does a schizophrenic see in hallucinations? & demons / spiders / Pokémon  \\ 
\hline
\end{tabular}
\end{adjustbox}
\end{table*}

\section{Discussion}
\label{sec:discussion}

Obtaining relevant, high-quality and privacy-compliant data remains a critical challenge in the development of medical question-answering (QA) systems~\cite{krishna2021hurdles,kell2024qa}. In particular, within the field of mental health and specifically schizophrenia, the complexity and subjectivity of individual experiences and symptomatology increase the need for authentic and nuanced patient-centered content. Traditional studies often rely heavily on structured clinical records or artificially constructed benchmark datasets, which may lack sufficient diversity, depth, and real-world relevance~\cite{mutabazi2021review,kell2024qa}. In contrast, this study introduces a novel approach that leverages web-scraped patient discussions from mental health forums, providing direct insights into patients' real-world concerns, experiences, and coping strategies. By capturing authentic and spontaneous patient-generated text, this approach substantially enhances the realism, coverage, and applicability of the resulting QA datasets. Specifically, the TGSM employed in this study effectively identifies core thematic aspects within the extensive corpus of user-generated posts. This method significantly improves systematic annotation of relevant QA pairs by clearly delineating the main topics and their corresponding questions and answers, thus increasing annotation efficiency, clarity, and semantic coherence. Compared to earlier studies using more conventional annotation methods, TGSM facilitates capturing multiple valid answers for individual questions, thus addressing critical open challenges identified in previous QA research, such as low question clarity and variability in annotation quality~\cite{alasmari2024quality}.

The dataset constructed through this approach comprises 415,602 user posts, systematically processed into 35 paragraphs, 35 distinct question types, and 1,050 carefully annotated answers. Integration of a retriever-reader pipeline further demonstrates the efficacy of using transformer-based language models (BioBERT, RoBERTa, and DistilBERT), achieving performance levels that match or exceed current state-of-the-art QA methods for mental health domains~\cite{c39,c14,c29}. Thus, this study showcases the practicality, scalability, and utility of leveraging authentic patient-generated forum data to improve QA model performance in schizophrenia-related contexts. 

Although our methodology can help ease some anxiety for patients with a particular disease by making it easier to find relevant information, several open critical challenges remain, offering promising avenues for future research. First, although public forums provide authentic patient-generated narratives, systematically ensuring the reliability, accuracy, and clinical relevance of such content remains an ongoing and unresolved challenge. Given the unstructured and informal nature of online discussions, further research into automated and semi-automated methods for validating and filtering clinically relevant content from patient-generated data could significantly enhance the quality and usability of forum-based QA datasets. In addition, expanding the methodological framework presented here to a broader spectrum of mental health conditions and patient demographics represents a valuable future research direction. Given the heterogeneity and complexity inherent in psychiatric disorders, systematically identifying suitable forums and replicating this framework across multiple mental health domains would enable the creation of larger and richer datasets, ultimately improving the generalizability and robustness of QA systems. Another important avenue involves exploring automated approaches to accelerate data collection and annotation. For example, recent advances in automated question generation, which involve generating contextually relevant and meaningful questions directly from given documents~\cite{c41}, present opportunities to enrich and diversify training datasets rapidly. Such methods can enhance efficiency, reduce manual annotation effort, and systematically expand dataset coverage, further improving model generalization and real-world applicability.

\nocite{zhang2026fairLMs,zhang2026fairAI,zhang2024ai,zhang2022longitudinal,zhang2023censored,zhang2019faht,yin2026disentangled,wang2025fgSmote,yin2025digitalForensicsLLM,wang2025redefiningFairness,wang2025unified_fair_graph_generation,wang2026towards,yin2024counterfactualSoftware,wang2025limitedDemographicsSurvey,yin2024fairAIED,wang2025fairnessAwareGraphLimitedDemo,yin2025amcr,amon2025uncertain,wang2025fairness,yin2026Toward,zhang2021disentangled,zhang2023individual}

\section{Conclusion}
\label{sec:conclusion}

In general, this work introduced a novel methodology for building a relevant medical text dataset and utilize it to build an efficient QA model to analyze the symptoms and impact of a particular disease on the daily lifestyle of an individual suffering from it. By identifying appropriate web forums for a specific disease, such as the ``Mental Health'' forum for individuals with schizophrenia and other mental health disorders, this study demonstrates how to construct a low-bias QA dataset with minimal privacy concerns in an optimized and accelerated manner. In addition, empirical results show that by fine-tuning the BioBERT model on our collected QA dataset, our proposed approach can outperform state-of-the-art models for mental health disorders. Our approach to creating QA datasets for more diseases efficiently, coupled with greater efforts to identify appropriate low-noise data sources for various diseases, can significantly help patients worldwide access reliable information about their disease.

\bibliography{main}

\end{document}